# FPGA Based Assembling of Facial Components for Human Face Construction

Santanu Halder[1], Debotosh Bhattacharjee[2], Mita Nasipuri[2], Dipak Kumar Basu[2*], Mahantapas Kundu[2]
[1]Department of Computer Science and Engineering, GNIT, Kolkata - 700114, India
Email: sant.halder@gmail.com
[2]Department of Computer Science and Engineering, Jadavpur University, Kolkata, 700032, India
Email: { debotosh@indiatimes.com, mita_nasipuri@gmail.com, dipakkbasu@gmail.com, mkundu@cse.jdvu.ac.in }
[*]AICTE Emeritus Fellow

*Abstract*-This paper aims at VLSI realization for generation of a new face from textual description. The FASY (FAce SYnthesis) System is a Face Database Retrieval and new Face generation System that is under development. One of its main features is the generation of the requested face when it is not found in the existing database. The new face generation system works in three steps – searching phase, assembling phase and tuning phase. In this paper the tuning phase using hardware description language and its implementation in a Field Programmable Gate Array (FPGA) device is presented.
*Index Terms*- Adjusting Intensities, Assembling the Components, Face Generation, VHDL, FPGA

## I. INTRODUCTION

Since early 1990's, Face Recognition Technology (FRT) becomes an active research area and there are a lot of works on facial recognition and facial feature extraction [1][2]. In 1994, Jian Kang Wu and Arcot Desai Narasimhalu [2] worked on "Identifying Faces Using Multiple Retrievals" where they retrieved the existing faces based on some properties of the different face components. But if the desired face is not in the database, then it is necessary to construct a new face from the user's description and this work is a new one which is one of the objective of the present work. This paper is a part of a main research effort, whose aim is the construction of the FASY-System based on human like description of each components of the face. The FASY System has the following features: (i) face queries using human-like description of the face, (ii) searching the required face in the database and (iii) generation of the requested face when they are not in the database. All the face components are stored in the database and are retrieved according to the user given queries and combined together to form the whole face which is then shown to the user. The Face Generation system works in 3 steps: (i) Searching phase to search the database for each single components of the face according to the user's query. (ii) Assembling phase to place all the components at proper places to generate the full face (iii) Tuning phase to adjust the intensities to cover up the stitching lines between different pair of face components so that the generated face looks natural. This paper is focused on the face generation system and the VLSI realization of the Tuning phase. Here we have worked with the databases found from AT & T Laboratories Cambridge [3]. The specifications for the database described in this paper are as follows:

- In all blank facial images, the ears should be visible.
- Sufficient variety of facial components and their properties should be stored in the database.
- The database should contain subject of different ethnicities, sample of both men and women and sample from different age groups.
- Variation due to head tilt, shift, rotation and scaling should be minimized as much as possible.

## II. FACE COMPONENTS AND ASSOCIATED PARAMETERS

The human-like face description that FASY accepts has been determined by a psychological study. As a result of this study, we found 7 face components and a set of associated parameters to be used in our system. The face components are Face cutting (with ears and hairs), right eyebrow, right eye, left eyebrow, left eye, nose and lips. The different parameters and their values for each of the seven components of a face are as follows:

**A. Face cutting**
(I)  Sex (Male, Female)
(II) Shape (Oval, Round, Cant Say)
(III) Hair Density (Highly Dense, Low Dense, Normal, Cant Say)

**B. Eyebrows (Left and Right)**
(I)   Length (Small, Large, Normal, Cant Say)
(II)  Width (Small, Large, Normal, Cant Say)
(III) Shape (Flat, Round, Wavy, Artistic, Cant Say)
(IV) Hair (Highly Dense, Low Dense, Normal, Cant Say)

**C. Eyes (Left and Right)**
(I)   Length (Small, Large, Normal, Cant Say)
(II)  Width (Small, Large, Normal, Cant Say)
(III) Shape (Round, Elliptic, Cant Say)

**D. Nose**
(I)   Sharpness (Sharp, Blunt, Normal, Cant Say)
(II)  Length (Small, Large, Normal, Cant Say)
(III) Width (Small, Large, Normal, Cant Say)

**E. Lip**
(I)   Length (Wide, Small, Normal, Cant Say)
(II)  Width (Thick, Thin, Normal, Cant Say)
(III) Shape (Linear, Wavy, Cant Say)

## III. THE FASY SYSTEM

A human user makes a query of a face using a human-like face description. The FD module interprets the query and translates it into a face description using different Face Parameters. This face description is used by the FR module to search the face in the database. The system retrieves the existing faces based on the user description. In the case when the desired face is not found in the database, the user can select the option for automatic generation of it through the Face Generator (FG) module which first finds the face components according to the user's description and generates a new face with the selected face parameters, which is presented to the user. If the user is not satisfied with the generated face, user can reset the query through the Parameter Adjustment (PA) module. Fig. 1 shows the block diagram of FASY system.

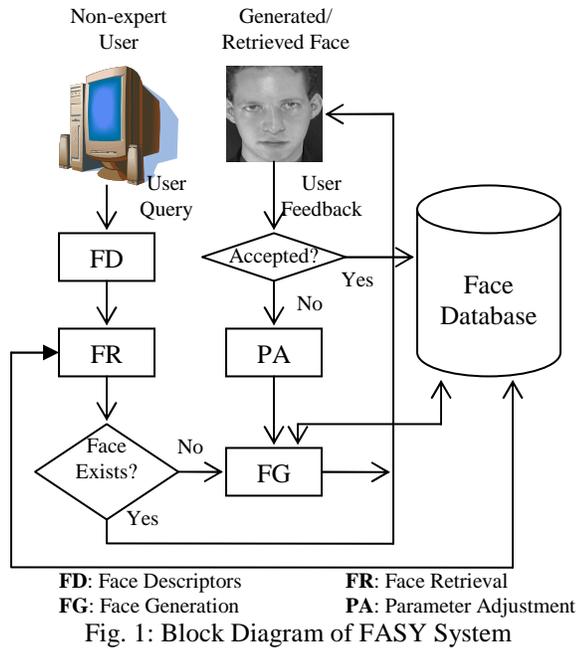

FD: Face Descriptors    FR: Face Retrieval
FG: Face Generation    PA: Parameter Adjustment
Fig. 1: Block Diagram of FASY System

## IV. SEARCHING PHASE

The searching phase accepts the user description of the different components of the face and searches the database for each single components of the face. The parameters for the facial cuttings and components are stored in the database previously. For a set of given parameters of a component multiple images may be selected and user can choose one of them.

## V. ASSEMBLING PHASE

The retrieved components of the face selected by the user based on user's query are fed to the assembling phase to combine all the components to generate the full face. The assembling phase accepts all the components retrieved from the database and place those components on the appropriate position of the face. The arrangement of the facial features does not change very much from person to person; therefore we can predict the position of each facial feature on a face when the position of the ear is known. To get the proper position of the components first we find the right ear position and this is done by the Algorithm 1.

*Algorithm 1*
Algorithm Find_Position _Ear(FaceI)
// FaceI is the intensity matrix of gray scale face.

1. {
2. Get the binary image of FaceI into a binary matrix *FaceBinary* with order m X n.
   // In binary image, 1 denotes white region and 0 denotes black region. So starting from first column find the first 1 value in FaceBinary matrix. Steps 6 to Step 14 do this.
3.    For j =1 to n
4.      For i = 1 to m
5.        If FaceBinary(i,j)=1
6.          Tx = i
7.          Ty = j
8.          Stop
9.        End if
10.      End for
11.    End for
12. }// End of Algorithm

Fig. 2 shows a face and Fig. 3 shows its binary image after intensity adjustment. Suppose $(T_x, T_y)$ is the co-ordinates of the upper left corner of the right ear. The next task is to track the position of the other components. In general, eyes lie on the almost same x co-ordinate of the upper left corner of the ear. Here the size of each faces is $92 \times 112$ and for this size we have assumed some constant values for calculating the positions of the components on the blank face. Hence the co-ordinates of the components for a normal face are calculated as follows by the experimental results:

Tx = Tx;
Ty = Ty +10;
*Position of the right eyebrow* = (Tx−5, Ty)
*Position of the right eye* = (Tx, Ty+ width of the right eyebrow−width of the right eye)
*Position of the nose* = (Tx−2, Ty+ width of the right eyebrow)
*Position of the left eyebrow* = (Tx−5, y- coordinate of nose + width of nose −5)
*Position of the left eye* = (Tx, y-coordinate of left eyebrow)
*Position of the lip*= (x-coordinate of nose + height of nose + 5, y-coordinate of nose + width of nose/2 − width of lip/2)

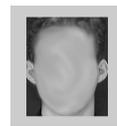 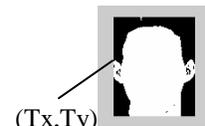

                                                    (Tx,Ty)
Fig. 2: Facial cutting        Fig. 3: Binary image

By the above procedure we find the normal positions of the face components. But for deformed shape of face these positions can vary and so this system has a provision to change the position of the components as required after tuning phase.

## VI. TUNING PHASE

The retrieved components when overlapped on the basic face cutting, the background intensity of the face may not match with the facial intensity. To make the overall face clear, prominent and natural to the user, our system works in two steps:
1. First get the binary image and from it find the actual region of the image.

2. Adjust the intensities so that the face looks natural.

Consider an image of an eye. Generally the shape of an eye is oval or round but dimension of the image of an eye as stored in the database is a rectangle which contain image of some face surface besides the eye. So instead of copying the whole image we can store the intensities of oval region of the eye portions only. For this purpose binary image of the components are used. In binary image, value 1 denotes the white region and value 0 denotes the black region. In Fig. 4, the black region contains the actual region of eye and eyebrow. So copy those pixel intensities of the eye image where the corresponding binary image contains 0.

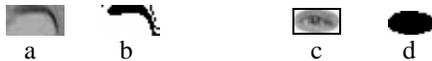

a　　　　b　　　　　　c　　　　d

Fig. 4: (a) Original left eyebrow (b) Binary image of left eyebrow after intensity adjustment (c) Original left eye (d) Binary image of left eye after intensity adjustment

Now to match the intensities of the different components with face cutting intensities we have designed an algorithm which works with the neighborhood intensities. Here we have considered 3 × 3 neighborhood intensities.

| $FI_{(x-1, y-1)}$ | $FI_{(x-1, y)}$ | $FI_{(x-1, y+1)}$ |
|---|---|---|
| $FI_{(x, y-1)}$ | $FI_{(x, y)}$ | $FI_{(x, y+1)}$ |
| $FI_{(x+1, y-1)}$ | $FI_{(x+1, y)}$ | $FI_{(x+1, y+1)}$ |

Fig. 5: 3 × 3 intensity matrix of face cutting for the (x, y) position

| $CI_{(i-1, j-1)}$ | $CI_{(i-1, j)}$ | $CI_{(i-1, j+1)}$ |
|---|---|---|
| $CI_{(i, j-1)}$ | $CI_{(i, j)}$ | $CI_{(i, j+1)}$ |
| $CI_{(i+1, j-1)}$ | $CI_{(i+1, j)}$ | $CI_{(i+1, j+1)}$ |

Fig. 6: 3 × 3 intensity matrix of a face component for the (i, j) position

Now instead of copying the original intensities of a component into the face cutting, we calculate a new set of intensities for each component as follows:

Suppose the intensity value of facial cutting at the position (x,y) is $FI_{(x,y)}$ and the intensity value of a face component at the position (i,j) is $CI_{(i,j)}$. $CI_{(i,j)}$ is to be copied at the (x,y) position on the face cutting. Now instead of replacing $FI_{(x,y)}$ by $CI_{(i,j)}$, we are calculating a new value of $CI_{(i,j)}$ for adjusting the intensity between face cutting and face components.

Summation of neighborhood intensities of face cutting at point (x,y) is

$F_I = FI_{(x-1, y-1)} + FI_{(x-1, y)} + FI_{(x-1, y+1)} + FI_{(x, y-1)} + FI_{(x, y)} + FI_{(x, y+1)} + FI_{(x+1, y-1)} + FI_{(x+1, y)} + FI_{(x+1, y+1)}$

Summation of neighborhood intensities of a face component at point (i,j) is-

$C_I = CI_{(i-1, j-1)} + CI_{(i-1, j)} + CI_{(i-1, j+1)} + CI_{(i, j-1)} + CI_{(i, j)} + CI_{(i, j+1)} + CI_{(i+1, j-1)} + CI_{(i+1, j)} + CI_{(i+1, j+1)}$

Intensity Factor $IF = F_I / C_I$

Now calculate the new intensity of face cutting at the point (x,y) as -

$FI_{(x, y)} = (FI_{(x, y)} + 2 * IF * CI_{(x, y)}) / (1 + 2 * IF)$

Algorithm 2 depicts the steps for tuning phase:

*Algorithm 2*

1. Algorithm Place_Image(I, F)
2. // I is intensity matrix of a face component with order m × n, $I_1$ is the binary matrix of the face component and F is the intensity matrix of facial cutting.
3. {
4. Get the binary image of I in the binary matrix $I_1$.
5. For each row i of the matrix I (i=1 to m)
6. For each column j of the matrix I (j=1 to n)
7. If $I_1(i, j)$ is 0 then
8. // Suppose intensity at (i, j) of the face component is to be copied at (x, y) position of the face cutting.
9. FI = F(x-1, y-1) + F(x-1, y) + F(x-1, y+1) + F(x, y-1) + F(x, y) + F(x, y+1) + F(x+1, y-1)+F(x+1, y) + F(x+1, y+1)
10. CI = I(i-1, j-1) + I(i-1, j) + I(i-1, j+1) + I(i, j-1) + I(i, j) + I(i, j+1)+ I(i+1, j-1) + I(i+1, j) + I(i+1, j+1)
11. IF = FI / CI
12. F(x, y) = (F(x, y) + 2 * IF * I(i, j)) / (1 + 2 * IF)
13. End if
14. End for
15. End for
16. } //End of Algorithm

Fig. 7 shows the steps to construct a new face with the selected face cutting and face components by the user. The position of the face components are calculated using the Algorithm 1 and the equations described in section V.

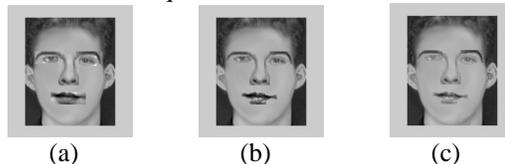

(a)　　　　　　(b)　　　　　　(c)

Fig. 7: Steps to generate a new face

(a)Blind replacement of the components on the face (b) Replacing the actual region of each component (c) Final image after intensity adjustment of the components (Applying Algorithm 2)

VII. HARDWARE IMPLEMENTATION OF TUNING PHASE

The high density techniques of current FPGAs will be highly attractive solution for our desired hardware implementation of software based new face generation algorithm as they provide flexibility, easy implementation and high quality performance along with less computational complexity. Moreover, FPGA has advantage of low investment cost and desktop testing with moderate processing speed thereby offering itself as suitable one for real time application. To reduce the hardware overhead, we have resized the facial image to the size of 23×28. For hardware implementation, first we find the value of (Tx,Ty) using Matlab-6.5. Then based on the value of (Tx,Ty), an image of Fig. 9 is constructed placing the selected facial components in their proper place on a 23 × 28 matrix using the method described in section V. Now the text files containing intensity values for the image of Fig. 8 and for the blank face are generated using Matlab-6.5 and stored in each separate text files. Algorithm 3 describes the steps for generation of a text file containing intensity values of an image.

Algorithm 3
Algorithm Generate_Text_File(I)
//I is the image for which the intensity values are to be stored in a text file named Image.txt.
1. {
2. Read the image I into the variable Image_Intensities.
3. Find the dimension of I. Suppose W stores the width and H contains the height of I.
4. Delete the file Image.txt.
5. Open the file Image.txt in append mode.
6. for i=1:H
7.    for j=1:W
8.       Print the value of Image_Intensities(i,j) into Image.txt.
9.    End For
10. End For
11. Close the file Image.txt.
12. }//End of Algorithm

Table 1 shows the name of the text files and their purposes.

Table 1: Name of the text files and their purposes

| Text file | Purpose |
|---|---|
| Face.txt | To store the intensity values of blank face. |
| Components.txt | To store the intensity values of lip. |

According to the method of tuning phase described in section VI, first the binary image of a component is required to find the actual region of an image. Suppose the threshold value T is set for this purpose. Now some changes are required on the Algorithm 2 as here we have already placed the components in their proper places before starting the cover-up of the stitching line. Algorithm 4 depicts the process of tuning phase for hardware implementation with the slight modification on Algorithm 2.

*Algorithm 4*
Algorithm Tuning_FPGA (I1, I2)
// I1 is the image of a blank face and I2 is the image of selected facial components which are placed in their proper places. The size of each image is m × n. T is the threshold value.
1. {
2. Copy the intensity values of I1 to I3
3. for x=2 to m
4.    for y=2 to n
5.       if I2(x,y) > T then
         j=x+1,k=y+1
6.         FI = $\sum_{\substack{j=x-1,k=y-1 \\ j=x+1,k=y+1}} I1(j,k)$
7.         CI = $\sum_{j=x-1,k=y-1} I2(j,k)$
8.         IF = FI / CI
9.         I3(x, y) = (I3(x, y) + 2 * IF * I2(i, j)) / (1 + 2 * IF)
10.       End If
11.    End For
12. End For
13. }//End of Algorithm

During the hardware realization, Simulink block level model is run with the integer values stored in the text files described in Table 1. All the VHDL codes are then simulated in Xilinx Webpack 4.1 for obtaining the RTL representations as well as technology views of the face generation method required for the hardware design. The RTL representations and technology views for the new face generation technique are given in Fig. 10 to Fig. 12.

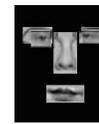

Fig. 8: A black face containing only the facial components in their proper place

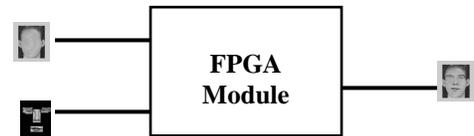

Fig. 9: The input and output images of the FPGA module

The analysis of the program is next reported:
DEVICE UTILIZATION SUMMARY:
SELECTED DEVICE                   : 2s15cs144-5
NUMBER OF SLICES                  : 41 OUT OF  192  21%
NUMBER OF SLICE FLIP FLOPS    : 72 OUT OF  384  18%
NUMBER OF 4 INPUT LUTs          : 41 OUT OF  384  10%
NUMBER OF GCLKs                   : 1 OUT OF    4   25%

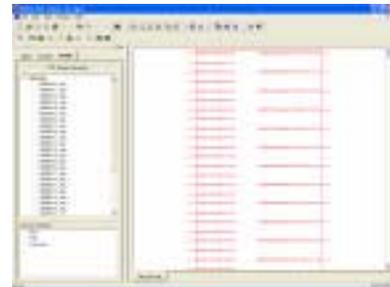

Fig. 10: RTL Representation of Top level – Tuning phase of new face construction in Webpack 4.1

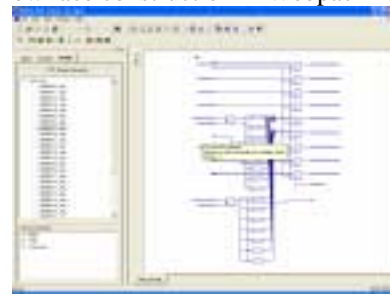

Fig. 11: RTL Representation(1) of Subsystem – Tuning phase of new face construction in Webpack 4.1

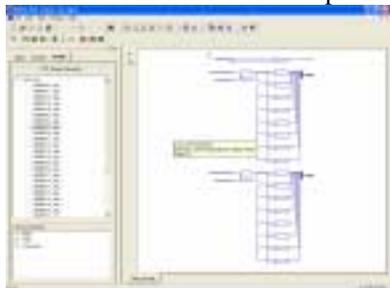

Fig. 12: RTL Representation (2) of Subsystem – Tuning phase of new face construction in Webpack 4.1

## VII. EXPERIMENTAL RESULTS

For testing the proposed method, we had about 200 male and female face images of different ages. From these images we obtained 100 gray scale facial cuttings, 50 gray scale left eyebrows, 50 gray scale left eyes, 50 gray scale right eyebrows, 50 gray scale right eyes, 30 gray scale noses and 30 gray scale lips. The different parameters for the facial cutting and components were set previously. In about 80% cases the FASY system generated the new faces successfully which fulfilled the user's query and looked natural. Suppose we want to generate a face which has the impressions described in Table 2:

Table 2: Face impressions of a person

| Components | Parameters | Values |
|---|---|---|
| Face Cutting | Sex | Male |
|  | Shape | Oval |
|  | Hair Density | Normal |
| Right Eyebrow | Length | Large |
|  | Width | Normal |
|  | Shape | Elliptic |
|  | Hair | Highly Dense |
| Right Eye | Length | Normal |
|  | Width | Normal |
|  | Shape | Elliptic |
| Left Eyebrow | Length | Large |
|  | Width | Normal |
|  | Shape | Elliptic |
|  | Hair | Highly Dense |
| Left Eye | Length | Normal |
|  | Width | Normal |
|  | Shape | Elliptic |
| Nose | Sharpness | Normal |
|  | Length | Normal |
|  | Width | Normal |
| Lips | Length | Normal |
|  | Width | Normal |
|  | Shape | Cant Say |

Fig. 13 shows the generated face according to the face impression of Table 2

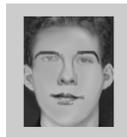

Fig. 13: Generated face according to the face impression of Table 2

Fig. 14 (b) and Fig. 15(b) show the constructed faces with the stored facial components in our database from the textual face impressions of two persons shown in Fig. 14(a) and Fig. 15(a) respectively.

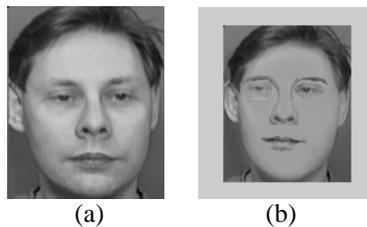

(a)         (b)

Fig 14: Generated face from the impression of the face given in Fig. 14(a)

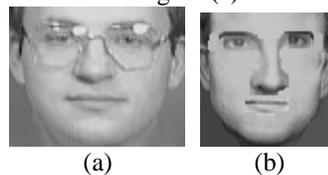

(a)         (b)

Fig. 15: Generated face from the impression of the face given in Fig. 15(a)

## VIII. CONCLUSION

Providing the facility of a user-friendly face generation system comes out to be a very complex task. The FASY system corresponds to an effort in this direction. The interactive retrieval process allows also an update of the face database. If the requested face can not be found in the database the user can choose to interactively generate it. The generation of the requested face when they are not found in the database is one of the main FASY characteristics. When the user interactively generates the face, the database can be expanded and the characteristics of the generated face that are supplied by the user are stored in the database as well. From this, even a small-at-the-beginning face database can grow from the very beginning on. This work can be especially useful for the criminal identification purpose where we have to construct a face of a criminal based on the description of eye-witness.

## REFERENCES

[1] Ming-Hsuan Yang, Kriegman, D.J., and Ahuja, N.; "Detecting faces in images: a survey," Pattern Analysis and Machine Intelligence, IEEE Transactions on Vol 24, Issue 1, pp34 - 58, Jan. 2002.

[2] Jian Kang Wu and Arcot Desai Narasimhalu, "Identifying Faces Using Multiple Retrievals", IEEE Multimedia, (ISSN 1070-986X), pp 27-38, Summer 1994

[3] http:// www.cl.cam.ac.UK/ Research/ DTG/ attarchive:pub/ data/ att_faces.zip.